\documentclass[journal]{IEEEtran}
\ifCLASSINFOpdf
\else
\fi

\usepackage{epsfig,graphicx}
\usepackage{amsmath,amssymb}
\usepackage{multirow,booktabs, hhline}
\usepackage{bm}
\usepackage{url}
\usepackage{cite}

\begin{document}
%

\title{Improving Sequence Modeling Ability of Recurrent Neural Networks via Sememes}
%
%
%

        
\author{Yujia~Qin, Fanchao~Qi, Sicong~Ouyang, Zhiyuan~Liu, Cheng~Yang,\\Yasheng~Wang, Qun~Liu, and Maosong~Sun%
\thanks{Y. Qin and F. Qi contribute equally to this work. This work was supported in part by the Natural Science Foundation of China (NSFC) and the German Research Foundation (DFG) in Project Crossmodal Learning, NSFC 61621136008 / DFG TRR-169, in part by the National Natural Science Foundation of China (NSFC No. 61732008), and in part by Beijing Academy of Artificial Intelligence (BAAI). \textit{(Corresponding author: Zhiyuan Liu.)}}%
\thanks{Y. Qin is with the Department of Electronic Engineering, Tsinghua University, Beijing 100084, China (e-mail: qinyj16@mails.tsinghua.edu.cn).}%
\thanks{F. Qi, Z. Liu and M. Sun are with the Department of Computer Science and Technology, Institute for Artificial Intelligence, Tsinghua University and Beijing National Research Center for Information Science and Technology, Beijing 100084, China (e-mail: qfc17@mails.tsinghua.edu.cn; liuzy@tsinghua.edu.cn; sms@tsinghua.edu.cn).}%
\thanks{S. Ouyang and C. Yang are with Beijing University of Posts and Telecommunications, Beijing 100876, China (e-mail: scouyang4354@gmail.com; albertyang33@gmail.com).}%
\thanks{Y. Wang and Q. Liu are with Huawei Noah's Ark Lab, Hong Kong, China (e-mail: wangyasheng@huawei.com; qun.liu@huawei.com).}%
}

%
%

\markboth{IEEE/ACM TRANSACTIONS ON AUDIO, SPEECH, AND LANGUAGE PROCESSING,~Vol.~xx, No.~x, August~xxxx}%
{Qin \MakeLowercase{\textit{et al.}}: Improving Sequence Modeling Ability of Recurrent Neural Networks via Sememes}
%



\maketitle

\begin{abstract}
Sememes, the minimum semantic units of human languages, have been successfully utilized in various natural language processing applications. 
However, most existing studies exploit sememes in specific tasks and few efforts are made to utilize sememes more fundamentally. 
In this paper, we propose to incorporate sememes into recurrent neural networks (RNNs) to improve their sequence modeling ability, which is beneficial to all kinds of downstream tasks. 
We design three different sememe incorporation methods and employ them in typical RNNs including LSTM, GRU and their bidirectional variants.
In evaluation, we use several benchmark datasets involving PTB and WikiText-2 for language modeling, SNLI for natural language inference and another two datasets for sentiment analysis and paraphrase detection.
Experimental results show evident and consistent improvement of our sememe-incorporated models compared with vanilla RNNs, which proves the effectiveness of our sememe incorporation methods. 
Moreover, we find the sememe-incorporated models have higher robustness and outperform adversarial training in defending adversarial attack.
All the code and data of this work can be obtained at \url{https://github.com/thunlp/SememeRNN}. 
\end{abstract}


%
\IEEEpeerreviewmaketitle

\section{Introduction}
\IEEEPARstart{A}{} word is the smallest unit of language, but its meaning can be split into smaller elements, i.e., \textit{sememes}.
For instance, the meaning of the word ``boy'' can be represented by the composition of meanings of ``human'', ``male'' and ``child'', while the meaning of ``girl'' can be represented by ``human'', ``female'' and ``child''.
In linguistics, a sememe is defined as the minimum unit of semantics \cite{bloomfield1926set}, which is atomic or indivisible.
Some linguists have the opinion that the meanings of all the words can be represented with a limited set of sememes, which is similar to the idea of \textit{semantic primitives} \cite{wierzbicka1996semantics}.
Considering sememes are usually implicit in words, researchers build sememe knowledge bases (KBs), which contain many words manually annotated with a set of predefined sememes, to utilize them. 
With the help of sememe KBs, sememes have been successfully applied to various natural language processing (NLP) tasks, e.g., word similarity computation \cite{liu2002word}, sentiment analysis \cite{fu2013multi}, word representation learning \cite{niu2017improved} and lexicon expansion \cite{zeng2018chinese}.

However, existing work usually exploits sememes for specific tasks and few efforts are made to utilize sememes in a more general and fundamental fashion.
\cite{gu2018language} make an attempt to incorporate sememes into a long short-term memory (LSTM) \cite{hochreiter1997long} language model to improve its performance. 
Nevertheless, their method uses sememes in the decoder step only and as a result, it is not applicable to other sequence modeling tasks.
To the best of our knowledge, no previous work tries to employ sememes to model better text sequences and achieve higher performance of downstream tasks.

In this paper, we propose to incorporate sememes into recurrent neural networks (RNNs) to improve their general sequence modeling ability, which is beneficial to all kinds of downstream NLP tasks. 
Some studies have tried to incorporate other linguistic knowledge into RNNs \cite{ahn2016neural,yang2017leveraging,parthasarathi2018extending,young2018augmenting}.
However, almost all of them utilize word-level KBs, which comprise relations between words, e.g., WordNet \cite{miller1998wordnet} and ConceptNet \cite{speer2013conceptnet}. 
Different from these KBs, sememe KBs use semantically infra-word elements (sememes) to compositionally explain meanings of words and focus on the relations between sememes and words. 
Therefore, it is difficult to directly adopt previous methods to incorporate sememes into RNNs.

To tackle this challenge, we specifically design three methods of incorporating sememes into RNNs.
All of them are highly adaptable and work on different RNN architecture.
We employ these methods in two typical RNNs including LSTM, gated recurrent unit (GRU) and their bidirectional variants.
In experiments, we evaluate the sememe-incorporated and vanilla RNNs on the benchmark datasets of several representative sequence modeling tasks, including language modeling, natural language inference, sentiment analysis and paraphrase detection.
Experimental results show that the sememe-incorporated RNNs achieve consistent and significant performance improvement on all the tasks compared with vanilla RNNs, which demonstrates the effectiveness of our sememe-incorporation methods and the usefulness of sememes.
We also make a case study to explain the benefit of sememes in the task of natural language inference.
Furthermore, we conduct an adversarial attack experiment, finding that the sememe-incorporated RNNs display higher robustness and perform much better than adversarial training when defending adversarial attack.

In conclusion, our contribution consists in: (1) making the first exploration of utilizing sememes to improve the general sequence modeling ability of RNNs; and (2) proposing three effective and highly adaptable methods of incorporating sememes into RNNs.

\section{Background}
In this section, we first introduce the sememe annotation in HowNet, the sememe KB we utilize. 
Then we give a brief introduction to two typical RNNs, namely LSTM and GRU, together with their bidirectional variants.


\subsection{Sememe Annotation in HowNet}

\begin{figure}[!t]
\centering
\includegraphics[width=0.95\linewidth]{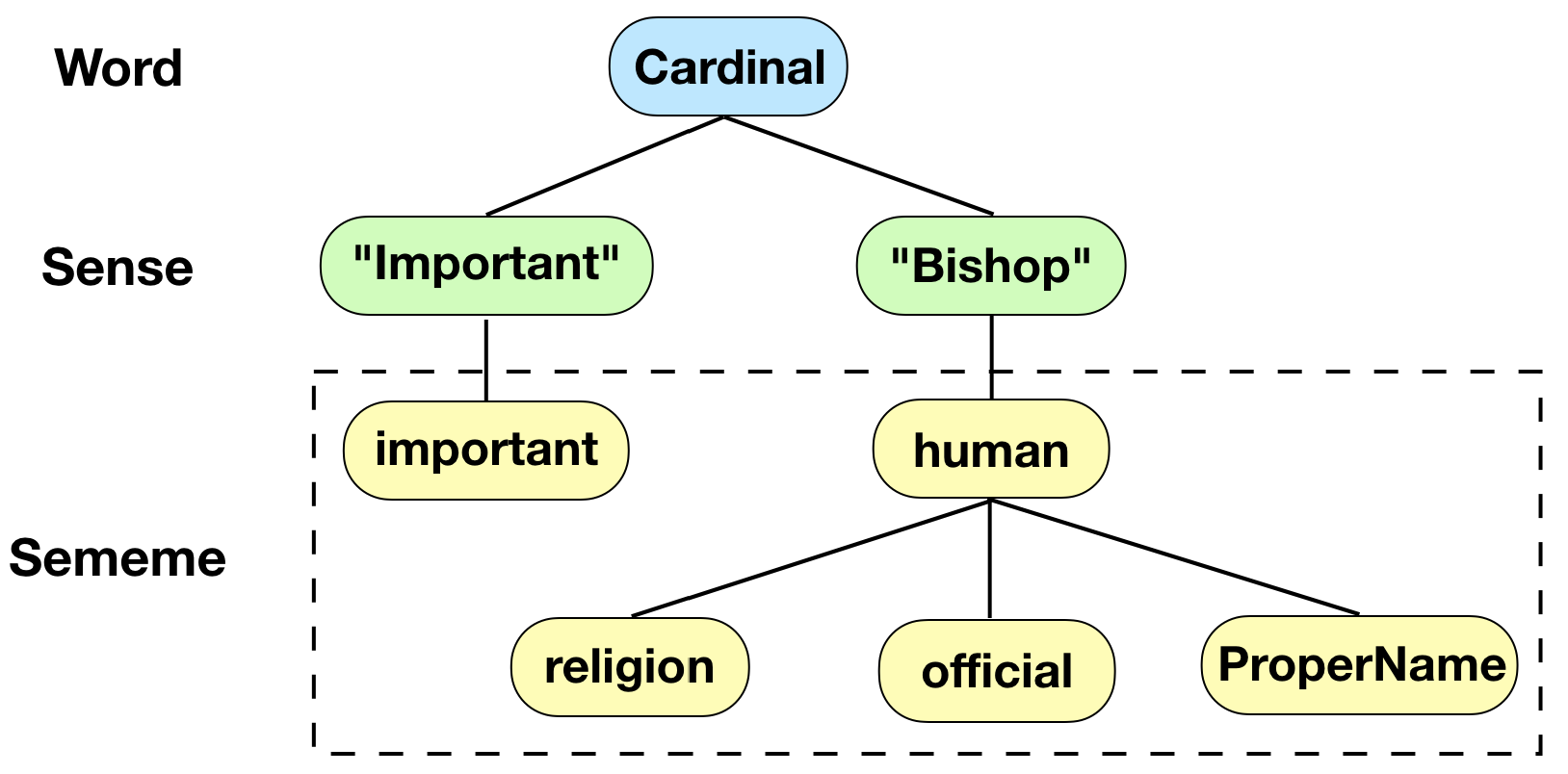}
\caption{{An example of how words are annotated with sememes in HowNet.}}
\label{fig:hownet}
\end{figure}

HowNet \cite{dong2003hownet} is one of the most famous sememe KBs, which contains over 100 thousand Chinese and English words annotated with about 2,000 predefined sememes. 
Sememe annotation in HowNet is sense-level.
In other words, each sense of polysemous words is annotated with one or more sememes with hierarchical structures. 
Fig. \ref{fig:hownet} illustrates the sememe annotation of the word ``cardinal'' in HowNet.
As shown in the figure, ``cardinal'' has two senses in HowNet, namely ``cardinal (important)'' and ``cardinal (bishop)''.
The former sense is annotated with only one sememe \texttt{important}, while the latter sense has one main sememe \texttt{human} and three subsidiary sememes including \texttt{religion}, \texttt{official} and \texttt{ProperName}. 

In this paper, we focus on the meanings of sememes and ignore their hierarchical structures for simplicity. 
Thus, we simply equip each word with a sememe set, which comprises all the sememes annotated to the senses of the word. 
For instance, the sememe set of ``cardinal'' is \{\texttt{important}, \texttt{human}, \texttt{religion}, \texttt{official}, \texttt{ProperName}\}.
We leave the utilization of sememe structures for future work.

\subsection{Introduction to Typical RNNs}
RNN is a class of artificial neural network designed for processing temporal sequences.
LSTM and GRU are two of the most predominant RNNs.
They have been widely used in recent years owing to their superior sequence modeling ability.

\begin{figure*}[t!]
\centering
\includegraphics[width=1.0\textwidth]{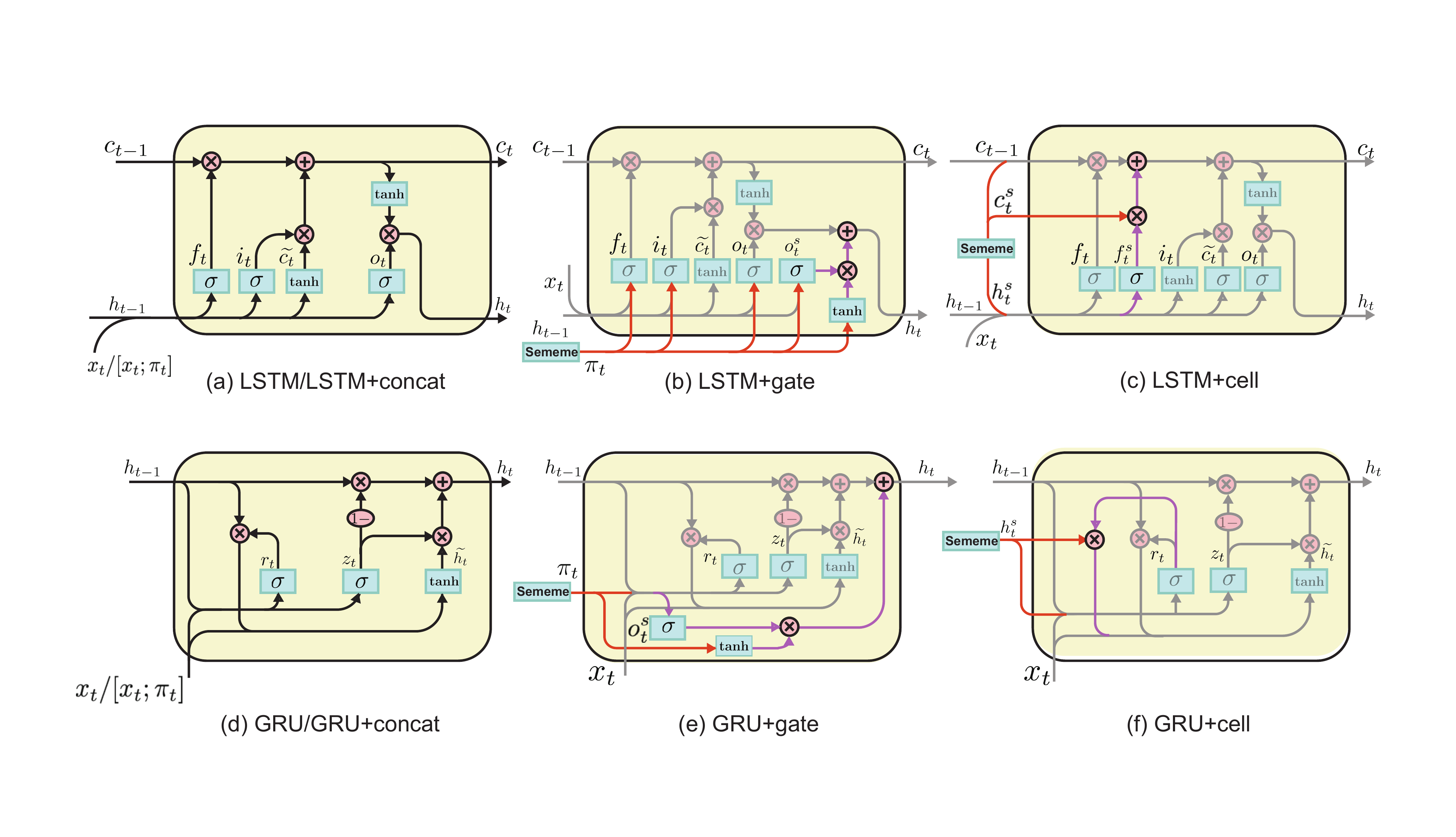}
\caption{The cell structures of three sememe-incorporated LSTMs and GRUs. (a), (b) and (c) illustrate the cell structures of vanilla LSTM (and with simple concatenation of sememe embeddings), LSTM with sememe output gate and LSTM with sememe-LSTM cell, respectively. 
(d), (e) and (f) exhibit the cell structures of vanilla GRU (and with simple concatenation of sememe embeddings), GRU with sememe output gate and GRU with sememe-GRU cell, respectively. 
In (b), (c), (e) and (f), the operators inherited from vanilla RNNs are in high transparency.
In addition, directly sememe-related information paths are denoted with red lines, and the other newly added information paths are denoted with purple lines.
}
\label{fig:intra}
\end{figure*}

An LSTM consists of multiple identical cells and each cell corresponds to a token in the input sequence. 
For each cell, it takes the embedding of the corresponding token $\mathbf{x}_t$ as input to update its cell state $\mathbf{c}_t$ and hidden state $\mathbf{h}_{t}$. 
Different from the basic RNN, LSTM integrates a forget gate $\mathbf{f}_t$, an input gate $\mathbf{i}_t$ and an output gate $\mathbf{o}_t$ into its cell, which can alleviate the gradient vanishing issue of the basic RNN. 
Given the hidden state $\mathbf{h}_{t-1}$ and the cell state $\mathbf{c}_{t-1}$ of the previous cell, the cell state $\mathbf{c}_t$ and the hidden state $\mathbf{h}_{t}$ of the current cell can be computed by:

\begin{equation}
\begin{aligned}
    \mathbf{f}_t &= \sigma(\mathbf{W}_f[\mathbf{x}_t; \mathbf{h}_{t-1}]+\mathbf{b}_f), \\
    \mathbf{i}_t &= \sigma(\mathbf{W}_I[\mathbf{x}_t; \mathbf{h}_{t-1}]+\mathbf{b}_I), \\
    \mathbf{\tilde{c}}_t &= \tanh(\mathbf{W}_c[\mathbf{x}_t; \mathbf{h}_{t-1}] + \mathbf{b}_c), \\
    \mathbf{c}_t &= \mathbf{f}_t * \mathbf{c}_{t-1} + \mathbf{i}_t * \mathbf{\tilde{c}}_t, \\
    \mathbf{o}_t &= \sigma(\mathbf{W}_o[\mathbf{x}_t; \mathbf{h}_{t-1}] + \mathbf{b}_o), \\
    \mathbf{h}_t &= \mathbf{o}_t * \tanh(\mathbf{c}_t), \\
\end{aligned}
\end{equation} 
where $\mathbf{W}_f$, $\mathbf{W}_I$, $\mathbf{W}_c$ and $\mathbf{W}_o$ are weight matrices, and $\mathbf{b}_f$, $\mathbf{b}_I$, $\mathbf{b}_c$ and $\mathbf{b}_o$ are bias vectors. 
$\sigma$ is the sigmoid function, $[\;]$ denotes the concatenation operation and $*$ indicates element wise multiplication.
The structure of an LSTM cell is illustrated in Fig. \ref{fig:intra}(a).


GRU is another popular extension of the basic RNN.
It has fewer gates than LSTM.
In addition to the input $\mathbf{x}_t$ and hidden state $\mathbf{h}_{t}$, each GRU cell embodies a update gate $\mathbf{z}_t$ and a reset gate $\mathbf{r}_t$.
The transition equations of GRU are as follows:
\begin{equation}
\begin{aligned}
    \mathbf{z}_t &= \sigma(\mathbf{W}_z[\mathbf{x}_t;\mathbf{h}_{t-1}] + \mathbf{b}_z),\\
    \mathbf{r}_t &= \sigma(\mathbf{W}_r[\mathbf{x}_t;\mathbf{h}_{t-1}] + \mathbf{b}_r),\\
    \mathbf{\tilde{h}}_t &= \tanh(\mathbf{W}_h[\mathbf{x}_t; \mathbf{r}_t * \mathbf{h}_{t-1}] + \mathbf{b}_h),\\
    \mathbf{h}_t &= (\mathbf{1}-\mathbf{z}_t) * \mathbf{h}_{t-1} + \mathbf{z}_t * \mathbf{\tilde{h}}_t,\\
\end{aligned}
\end{equation}
where $\mathbf{W}_z$, $\mathbf{W}_r$  $\mathbf{W}_h$ are weight matrices, and $\mathbf{b}_z$, $\mathbf{b}_r$, $\mathbf{b}_h$ are bias vectors. 
Fig. \ref{fig:intra}(d) shows the structure of a GRU cell.

Since both LSTM and GRU can only process sequences unidirectionally, their bidirectional variants, namely BiLSTM and BiGRU, are proposed to eliminate the restriction. 
BiLSTM and BiGRU have two sequences of cells: one processes the input sequence from left to right and the other from right to left. 
Hence, each token in the input sequence corresponds two unidirectional hidden states, which are concatenated into bidirectional hidden states:
 \begin{equation}
   \begin{aligned}
\left[\overleftrightarrow{\mathbf{h}_1}, \overleftrightarrow{\mathbf{h}_2},..., \overleftrightarrow{\mathbf{h}_T}\right] &= \left[ \begin{array}{lr}
          \overrightarrow{\mathbf{h}_1}, \overrightarrow{\mathbf{h}_2},...,\overrightarrow{\mathbf{h}_T}, \\
          \overleftarrow{\mathbf{h}_1}, \overleftarrow{\mathbf{h}_2},...,\overleftarrow{\mathbf{h}_T}, \\
             \end{array}
             \right],
   \end{aligned}
  \end{equation}
where $T$ denotes the length of the input sequence.

\section{Sememe Incorporation Methods}

In this section, we elaborately describe three methods of incorporating sememes into RNNs, namely simple concatenation (+concat), adding sememe output gate (+gate) and introducing sememe-RNN cell (+cell).
All the three methods are applicable to both LSTM and GRU, both unidirectional and bidirectional RNNs. 
The following descriptions are based on unidirectional LSTM and GRU, and the newly added variables and formulae are underlined for clarity. 
Bidirectional sememe-incorporated LSTMs and GRUs are just a trivial matter of concatenating the hidden states of their unidirectional versions.
The cell structures of three different sememe-incorporated LSTMs and GRUs are exhibited in Fig. \ref{fig:intra}.


\subsection{Simple Concatenation}
 
The first sememe incorporation approach is quite straightforward.
It simply concatenates the sum of sememe embeddings of a word with the corresponding word embedding. 
In this way, sememe knowledge is incorporated into a RNN cell via its input (word embedding).
Formally, 
given a word $x_t$, its sememe set is $\mathcal{S}_t=\{s_1, \cdots, s_{|\mathcal{S}_t|}\}$, where $|\cdot|$ denotes the cardinality of a set.
We concatenate its original word embedding $\mathbf{x}_t$ with sememe knowledge embedding $\bm{\pi}_t$:
\begin{equation}
\begin{aligned}
    \bm{\pi}_t &= \frac{1}{|\mathcal{S}_t|} \sum_{s\in \mathcal{S}_t} \mathbf{s} \quad
    \mathbf{\tilde{x}}_t &= [\mathbf{x}_t; \bm{\pi}_t],\\
\end{aligned}
\end{equation}
where $\mathbf{s}$ denotes the sememe embedding of $s$, and $\mathbf{\tilde{x}}_t$ is the retrofitted word embedding which incorporates sememe knowledge.

\subsection{Adding Sememe Output Gate}

In the first method, sememe knowledge is shallowly incorporated into RNN cells.
It is essentially a kind of enhancement of word embeddings. 
Inspired by \cite{ma2018targeted}, we propose the second method which enables deep incorporation of sememes into RNN cells. 
More specifically, we add an additional sememe output gate $\mathbf{o}_t^s$ to the vanilla RNN cell, which decides how much sememe knowledge is absorbed into the hidden state. 
Meanwhile, the original gates are also affected by the sememe knowledge term. 
The transition equations of LSTM with the sememe output gate are as follows:
\begin{equation}
\begin{aligned}
\label{eq:LSTM2}
\mathbf{f}_t &= \sigma(\mathbf{W}_f[\mathbf{x}_t; \mathbf{h}_{t-1}; \underline{\bm{\pi}_t} ]+\mathbf{b}_f), \\
\mathbf{i}_t &= \sigma(\mathbf{W}_I[\mathbf{x}_t; \mathbf{h}_{t-1}; \underline{\bm{\pi}_t}]+\mathbf{b}_i), \\
\mathbf{\tilde{c}}_t &= \tanh(\mathbf{W}_c[\mathbf{x}_t; \mathbf{h}_{t-1}] + \mathbf{b}_c), \\
\mathbf{c}_t &= \mathbf{f}_t * \mathbf{c}_{t-1} + \mathbf{i}_t * \mathbf{\tilde{c}}_t, \\
\mathbf{o}_t &= \sigma(\mathbf{W}_o[\mathbf{x}_t; \mathbf{h}_{t-1}; \underline{\bm{\pi}_t} ] + \mathbf{b}_o), \\
\underline{\mathbf{o}_t^s} &  = \underline{ \sigma(\mathbf{W}_{o^s}[\mathbf{x}_t; \mathbf{h}_{t-1}; \bm{\pi}_t ] + \mathbf{b}_{o^s})}, \\
\mathbf{h}_t &= \mathbf{o}_t * \tanh(\mathbf{c}_t) + \underline{\mathbf{o}_t^s * \tanh(\mathbf{W}_c \bm{\pi}_t)},\\
\end{aligned}
\end{equation}
where $\mathbf{W}_{o^s}$ is a weight matrix and $\mathbf{b}_{o^s}$ is a bias vector.

We can find that $\underline{\mathbf{o}_t^s * \tanh(\mathbf{W}_c \bm{\pi}_t) }$ can directly add the term of sememe knowledge to the original hidden state.
By doing this, the original hidden state, which carries contextual information only, is enhanced by semantic information of sememe knowledge.

The sememe output gate can be added to GRU in a similar way, and the transition equations are as follows:
\begin{equation}
\begin{aligned}
\mathbf{z}_t &= \sigma(\mathbf{W}_z[\mathbf{x}_t; \mathbf{h}_{t-1}; \underline{\bm{\pi}_t }] + \mathbf{b}_z),\\
\mathbf{r}_t &= \sigma(\mathbf{W}_r[\mathbf{x}_t; \mathbf{h}_{t-1}; \underline{\bm{\pi}_t}] + \mathbf{b}_r),\\
\underline{\mathbf{o}^s_t} & = \underline{\sigma(\mathbf{W}_o[\mathbf{x}_t; \mathbf{h}_{t-1}; \bm{\pi}_t] + \mathbf{b}_o)}, \\
\mathbf{\tilde{h}}_t &= \tanh(\mathbf{W}_h[\mathbf{x}_t; \mathbf{r}_t * \mathbf{h}_{t-1}] + \mathbf{b}_h),\\
\mathbf{h}_t &= (\mathbf{1}-\mathbf{z}_t) * \mathbf{h}_{t-1} + \mathbf{z}_t * \mathbf{\tilde{h}}_t + \underline{\mathbf{o}^s_t \tanh(\bm{\pi}_t) },\\
\end{aligned}
\end{equation} 
where $\mathbf{o}^s_t$ is the sememe output gate, $\mathbf{W}_{o}$ is a weight matrix and $\mathbf{b}_{o}$ is a bias vector.


\subsection{Introducing Sememe-RNN Cell}
In the second method, although sememe knowledge is incorporated into RNN cells directly and deeply, it can be utilized more sufficiently.
Taking LSTM for example, as shown in Equation \eqref{eq:LSTM2}, the hidden state $\mathbf{h}_t$ comprises two parts. 
The first part is the contextual information item $\mathbf{o}_t * \tanh(\mathbf{c}_t)$. 
It bears the information of current word and preceding text, which has been processed by the forget gate and encoded into the cell state.
The second part is the sememe knowledge item $\underline{ \mathbf{o}_t^s *\tanh(\mathbf{W}_c \bm{\pi}_t) }$. 
It carries the information of sememe knowledge which has not processed or encoded.
Therefore, the two parts are inconsistent.


To address the issue, we propose the third sememe incorporation method. 
We regard the sememe knowledge as another information source like the previous word and introduce an extra RNN cell to encode it. 
Specifically, we first feed the sememe knowledge embedding to an LSTM cell (sememe-LSTM cell) and obtain its cell and hidden states which carry sememe knowledge.
Then we design a special forget gate for the sememe knowledge and use it to process the cell state of the sememe-LSTM cell, just as the previous cell state.
Finally, we add the processed cell state of the sememe-LSTM to the original cell state.
In addition, the hidden state of the sememe-LSTM cell is also absorbed into the input gate and output gate.
Formally, the transition equations of LSTM with the sememe-LSTM cell are as follows:
\begin{equation}
\begin{aligned}
\underline{{\mathbf{c}_t^s}} & \underline{,{\mathbf{h}_t^s}} = \underline{{\rm LSTM}^{(S)}(\bm{\pi}_t) }, \\
\mathbf{f}_t &= \sigma(\mathbf{W}_f[\mathbf{x}_t; \mathbf{h}_{t-1}
]+\mathbf{b}_f), \\
\underline{\mathbf{f}_{t}^s} &= \underline{\sigma(\mathbf{W}_f^s[\mathbf{x}_t; 
\mathbf{h}_t^s]+\mathbf{b}_f^s)}, \\
\mathbf{i}_t &= \sigma(\mathbf{W}_I[\mathbf{x}_t; \mathbf{h}_{t-1}; \underline{\mathbf{h}_t^s }]+\mathbf{b}_i), \\
\mathbf{\tilde{c}}_t &= \tanh(\mathbf{W}_c[\mathbf{x}_t; \mathbf{h}_{t-1}; \underline{\mathbf{h}_t^s }] + \mathbf{b}_c), \\
\mathbf{o}_t &= \sigma(\mathbf{W}_o[\mathbf{x}_t; \mathbf{h}_{t-1}; \underline{\mathbf{h}_t^s }] + \mathbf{b}_o), \\
\mathbf{c}_t &= \mathbf{f}_t * \mathbf{c}_{t-1} + \underline{\mathbf{f}_t^s * \mathbf{c}_t^s } + \mathbf{i}_t * \mathbf{\tilde{c}}_t, \\
\mathbf{h}_t &= \mathbf{o}_t * \tanh(\mathbf{c}_t), \\
\end{aligned}
\end{equation} 
where $\mathbf{c}_t^s$ and $\mathbf{h}_t^s$ are the cell state and hidden state of the sememe-LSTM cell, $\mathbf{f}_t^s$ is the sememe forget gate, $\mathbf{W}_{f^s}$ is a weight matrix and $\mathbf{b}_{f^s}$ is a bias vector.

Similarly, we can introduce a sememe-GRU cell to the original GRU and the transition equations are as follows:
 \begin{equation}
\label{eq:GRU+cell}
  \begin{aligned}
\underline{{\mathbf{h}_t^s}} &= \underline{GRU^{(S)}(\bm{\pi}_t)}, \\
\mathbf{z}_t &= \sigma(\mathbf{W}_z[\mathbf{x}_t; \mathbf{h}_{t-1}; \underline{\mathbf{h}_t^s }] + \mathbf{b}_z), \\
\mathbf{r}_t &= \sigma(\mathbf{W}_r[\mathbf{x}_t; \mathbf{h}_{t-1}; \underline{\mathbf{h}_t^s }] + \mathbf{b}_r), \\
\mathbf{\tilde{h}}_t &= \tanh(\mathbf{W}_h[\mathbf{x}_t; \mathbf{r}_t * (\mathbf{h}_{t-1} + \underline{\mathbf{h}_t^s)}] + \mathbf{b}_h),\\
\mathbf{h}_t &= (\mathbf{1}-\mathbf{z}_t) * \mathbf{h}_{t-1} + \mathbf{z}_t * \mathbf{\tilde{h}}_t,\\
\end{aligned}
\end{equation} 
where $\mathbf{h}_t^s$ is the hidden state of the sememe-GRU cell.

\section{Language Modeling}
In this section, we evaluate our sememe-incorporated RNNs on the task of language modeling (LM). 

\subsection{Dataset} 
We use two benchmark LM datasets, namely Penn Treebank (PTB) \cite{marcus1993building} and WikiText-2 \cite{merity2017pointer}.
PTB is made up of articles from the Wall Street Journal.
Its vocabulary size is $10,000$. 
The token numbers of its training, validation and test sets are $887,521$, $70,390$ and $78,669$ respectively.
WikiText-2 comprises Wikipedia articles and its vocabulary size is $33,278$.
It has $2,088,628$, $217,646$ and $245,569$	tokens in its training, validation and test sets.

We choose HowNet as the source of sememes. 
It contains $2,186$ different sememes and $43,321$ English words with sememe annotation.
We use the open-source API of HowNet, OpenHowNet \cite{qi2019openhownet}, to obtain annotated sememes of a word.
The numbers of sememe-annotated tokens in PTB and WikiText-2 are $870,520$ ($83.98\%$) and $2,068,779$ ($81.07\%$) respectively. 
For the words without sememe annotations, we simply set their sememe knowledge embeddings to $\mathbf{0}$.
 
\subsection{Experimental Settings}
\paragraph{Baseline Methods}
We choose the vanilla LSTM and GRU as the baseline methods. 
Notice that bidirectional RNNs are generally not used in the LM task because they are not allowed to know the whole sentence.

\paragraph{Hyper-parameters}
Following previous work, we try the models on two sets of hyper-parameters, namely ``medium'' and ``large''. 
For ``medium'', the dimension of hidden states and word/sememe embeddings is set to 650, the batch size is 20, and the dropout rate is 0.5. 
For ``large'', the dimension of vectors is 1500, the dropout rate is 0.65, and other hyper-parameters are the same as ``medium''.
All the word and sememe embeddings are randomly initialized as real-valued vectors using a normal distribution with mean 0 and variance 0.05.
The above-mentioned hyper-parameter settings are applied to all the models.

\paragraph{Training Strategy}
We adopt the same training strategy for all the models.
We choose stochastic gradient descent (SGD) as the optimizer, whose initial learning rate is 20 for LSTM and 10 for GRU. The learning rate would be divided by 4 if no improvement is observed on the validation set. 
The maximum training epoch number is 40 and the gradient norm clip boundary is 0.25.


\begin{table}[!t]
  \centering
  \small
    \caption{Perplexity Results of All the Models on the Validation and Test Sets of PTB and WikiText-2. (*) Denotes p$<$0.05 on a two-tailed t-test, against the vanilla models.}
    \begin{tabular}{lrrrr}
    \toprule
    Dataset & \multicolumn{2}{c}{PTB}&\multicolumn{2}{c}{WikiText-2} \\
    \cmidrule{1-5} 
    Model & {Valid} & {Test} & {Valid} & {Test} \\
    \midrule
    LSTM(medium)  & 84.48 & 80.86 & 99.31 & 93.88 \\
    \quad+concat & 81.79 & 78.90 & 96.05 & 91.41 \\
    \quad+gate & 81.15 & 77.73  & 95.27  & 90.19  \\
    \quad+cell(*) & {\textbf{79.67}} & \textbf{76.65} & \textbf{94.49} & \textbf{89.16} \\
    \hline
    LSTM(large)  & 80.63 & 77.34 & 96.25 & 90.77 \\
    \quad+concat & 78.35 & 75.25 & 92.72 & 87.51 \\
    \quad+gate & 77.02 & 73.90 & 91.02 & 86.16 \\    
    \quad+cell(*) & \textbf{76.15} & \textbf{73.87} & \textbf{90.52} & \textbf{85.76} \\
    \midrule
    GRU(medium)   & 94.01 & 90.68 & 109.38 & 103.04 \\
    \quad+concat & 90.68 & 87.29 & 105.11 & 98.89 \\
    \quad+gate(*) & \textbf{88.21} & \textbf{84.35} & \textbf{103.37} & \textbf{97.54} \\
    \quad+cell & 89.56 & 86.49 & 103.53 & 97.65 \\
    \hline
    GRU(large)   & 92.89 & 89.57  & 108.28 & 101.77 \\
    \quad+concat & 91.64 & 87.62 & 104.33 & 98.15 \\
    \quad+gate(*) & \textbf{88.01} & \textbf{84.60} & 102.07 & 96.11 \\
    \quad+cell & 89.33 & 86.05 & \textbf{101.22} & \textbf{95.67} \\
    \bottomrule
    \end{tabular}%
  \label{tab:lm}%
\end{table}%


\subsection{Experimental Results}
Table \ref{tab:lm} shows the perplexity results on both validation and test sets of the two datasets.
From the table, we can observe that: 

(1) All the sememe-incorporated RNNs, including the simplest +concat models, achieve lower perplexity as compared to corresponding vanilla RNNs,
which demonstrates the usefulness of sememes for enhancing sequence modeling and the effectiveness of our sememe incorporation methods;

(2) Among the three different methods, +cell performs best for LSTM and +gate performs best for GRU at both ``medium'' and ``large'' hyper-parameter settings. 
The possible explanation is that +cell incorporates limited sememe knowledge into GRU as compared to LSTM. 
In fact, GRU+cell is much less affected by sememe knowledge than GRU+gate, as shown in Equation \ref{eq:GRU+cell}.

(3) By comparing the results of the ``large'' vanilla RNNs and the ``medium'' sememe-incorporated RNNs, we can exclude the possibility that better performance of the sememe-incorporated RNNs is brought by more parameters.
For example, the perplexity of the ``large'' vanilla LSTM and the ``medium'' LSTM+cell on the four sets is 81.88/79.71, 78.34/76.57, 96.86/94.49 and 91.07/89.39 respectively. 
Although the ``large'' vanilla LSTM has much more parameters than ``medium'' LSTM+cell (76M vs. 24M), it is still outperformed by the latter.



\section{Natural Language Inference}
Natural language inference (NLI), also known as recognizing textual entailment, is a classic sentence pair classification task. 
In this section, we evaluate our models on the NLI task.

\subsection{Dataset}
We choose the most famous benchmark dataset, Stanford Natural Language Inference (SNLI) \cite{bowman2015alarge} for evaluation.
It contains 570k sentence pairs, each of which comprises a premise and a hypothesis.
The relations of the sentence pairs are manually classified into 3 categories, namely ``entailment'', ``contradiction'' and ``neutral''. 
The coverage of its sememe-annotated tokens is $82.31\%$.
For the words without sememe annotations, we still set the corresponding sememe knowledge embeddings to $\mathbf{0}$.

\subsection{Experimental Settings}

\paragraph{Baseline Methods}
We choose vanilla LSTM, GRU and their bidirectional variants (BiLSTM and BiGRU) as baseline methods.

\paragraph{Hyper-parameters}
For the input to the models, we use 300-dimensional word embeddings pre-trained by GloVe \cite{pennington2014glove}.
Besides, we also try concatenating GloVe embeddings with 256-dimensional ELMo embeddings \cite{peters2018deep} and BERT embeddings \cite{devlin2019bert} respectively, both of which capture contextual information.
These embeddings are frozen during training. 
The dropout rate for input word embedding is 0.2.
The dimension of hidden states is 2048.
In addition, other hyper-parameters are the same as those of the LM experiment.

\paragraph{Training Strategy}
We still choose the SGD optimizer, whose initial learning rate is 0.1 and weight factor is 0.99.
We divide the learning rate by 5 if no improvement is observed on the validation dataset. 

\paragraph{Classifier}
Following previous work \cite{bowman2016fast,mou2016natural}, we employ a three-layer perceptron plus a three-way softmax layer as the classifier, whose input is a feature vector constructed from the embeddings of a pair of sentences. 
Specifically, we use any RNN model to process the two sentences of a premise-hypothesis pair and obtain their embeddings $\mathbf{h}_{pre}$ and $\mathbf{h}_{hyp}$.
Then we construct the feature vector $\mathbf{v}$ as follows: 

\begin{equation}
\begin{aligned}
\centering
\mathbf{v} = \left[ \begin{array}{cc}
          \mathbf{h}_{pre} \\
          \mathbf{h}_{hyp} \\
          |\mathbf{h}_{pre} - \mathbf{h}_{hyp}| \\
          \mathbf{h}_{pre} * \mathbf{h}_{hyp}
             \end{array}
             \right].
\end{aligned}
\end{equation}

\begin{table}[!tbp]
  \small
  \centering
     \caption{Accuracy Results of All the Models on SNLI. (*) Denotes p$<$0.05 on a two-tailed t-test, against the vanilla models.}
    \begin{tabular}{clrrrr}
    \toprule
    Embedding & Model & {LSTM} & {GRU} & {BiLSTM} & {BiGRU} \\
    \midrule
    \multirow{4}{*}{GloVe} & vanilla & 81.03 & 81.59 & 81.38 & 81.92 \\
    & \ +concat & \textbf{81.72} & 81.87 & 82.39  & 82.70 \\
    & \ +gate & 81.61 & 82.33 & {83.06} & 83.18 \\
    & \ +cell(*) & {81.66} & \textbf{82.78} & \textbf{83.67} & \textbf{83.26} \\
    \midrule
    \multirow{4}{*}{GloVe+ELMo} & vanilla & 81.99 & 82.49 & 83.04 & 83.40\\
    & \ +concat & 82.47 & 81.98 & 83.23 & 83.23\\
    & \ +gate & 82.24 & \textbf{82.75} & 83.44 & 83.59\\
    & \ +cell(*) & \textbf{82.54} & 82.60 & \textbf{83.88} & \textbf{84.11}\\
    \midrule
    \multirow{4}{*}{GloVe+BERT} & vanilla & 82.26 & 82.66 & 83.16 & 83.45\\
    & \ +concat & {82.68} & 82.12 & 83.38& 83.51\\
    & \ +gate & 82.39 & \textbf{82.93} & 83.72 & 83.82\\
    & \ +cell(*) & \textbf{82.74} & 82.69 & \textbf{84.35} & \textbf{84.42}\\
    \bottomrule
    \end{tabular}%
  \label{tab:snli}%
\end{table}%


\subsection{Experimental Results}

Table \ref{tab:snli} lists the results of all the models on the test set of SNLI. 
From this table, we can see that: 

(1) All the sememe-incorporated models achieve marked performance enhancement compared with corresponding vanilla models, which proves the usefulness of sememes in improving the sentence representation ability of RNNs and the effectiveness of our sememe incorporation methods;


(2) Among the three sememe incorporation methods, +cell achieves the best overall performance, which manifests the great efficiency of +cell in utilizing sememe knowledge.
This is also consistent with the conclusion of the LM experiment basically.

\begin{table}[!t]
\small
\centering
\caption{Two Examples of Premise-hypothesis Pairs in the SNLI dataset. The Words In Italic Type are Important Words and their Sememes are Appended.}
\begin{tabular}{p{7.5cm}}
  \toprule
  {Entailment Example}\\
\textbf{Premise}: Four men stand in a circle facing each other \textit{playing} [\texttt{perform}, \texttt{reaction}, \texttt{MusicTool}] brass \textit{instruments} [\texttt{MusicTool}, \texttt{implement}] which people watch them.\\
  \textbf{Hypothesis}: The men are playing \textit{music} [\texttt{music}]. \\
  \midrule
  {Contradict Example}\\
  \textbf{Premise}: A group of women playing volleyball \textit{indoors} [\texttt{location}, \texttt{house}, \texttt{internal}].\\
  \textbf{Hypothesis}: People are \textit{outside} [\texttt{location}, \texttt{external}] tossing a ball.\\
  \bottomrule
\end{tabular}
\label{tb:nli_case}
\end{table}	

\subsection{Case Study}
In this subsection, we use two examples to illustrate how sememes are beneficial to handling the task of NLI.
Table \ref{tb:nli_case} exhibits two premise-hypothesis pairs in the SNLI dataset, where the sememes of some important words are appended to corresponding words.

For the first premise-hypothesis pair, its relation type is annotated as ``entailment'' in SNLI. 
Our sememe-incorporated models yield the correct result while all the baseline methods do not.
We notice that there are several important words whose sememes provide useful information. 
In the premise, both the words ``playing'' and ``instruments'' have the sememe \texttt{MusicTool}, which is semantically related to the sememe \texttt{music} of the word ``music'' in the hypothesis. 
We speculate that the semantic relatedness given by sememes assists our models in coping with this sentence pair.

The second premise-hypothesis pair, whose true relation is ``contradict'', is also classified correctly by our models but wrongly by the baseline methods.
We find that the word ``indoors'' in the premise has the sememe \texttt{internal}, while the word ``outside'' in the hypothesis has the sememe \texttt{external}. 
The two sememes are a pair of antonyms, which may explain why our models make the right judgment.

\section{Text Classification}
In this section, we evaluate our sememe-incorporated RNNs on two text classification tasks, namely sentiment analysis and paraphrase detection.

\subsection{Dataset}
For sentiment analysis, we use the CR dataset \cite{hu2004mining}. It contains about 8k product reviews (4k in the training set and 4k in the test set) and each review is labeled with ``positive'' or ``negative''.
For paraphrase detection, we choose the Microsoft Research Paraphrase Corpus (MRPC) \cite{dolan2004unsupervised}. It comprised pairs of sentences extracted from news sources on the Web. Each sentence pair is human-annotated according to whether it captures a paraphrase/semantic equivalence relationship.
It has 4.1k and 1.7k sentence pairs in its training and test sets.

\subsection{Experimental Settings}
Following previous work \cite{conneau2018senteval}, we transfer all the models trained on SNLI to new datasets.
Specifically, we use the datasets of the two tasks separately to continue to train the models which have already been trained on SNLI.
The settings of hyper-parameters and training are the same as those in \cite{conneau2018senteval}.

\begin{table}[!tbp]
\small
\centering
\caption{Accuracy Results of All the Models on CR and MRPC. (*) Denotes p$<$0.05 on a two-tailed t-test, against the vanilla models.}
    \begin{tabular}{clrrrr}
    \toprule
    Dataset & Model & {LSTM} & {GRU} & {BiLSTM} & {BiGRU}\\
    \midrule
    \multirow{4}{*}{CR}  & vanilla & 76.03 & 76.02 & 75.81 & 75.64 \\
    & \ +concat & 76.94 & 77.38 & \textbf{78.06}  & 76.28 \\
    & \ +gate & \textbf{77.52} & 77.95 & 77.15 & \textbf{77.50} \\
    & \ +cell(*) & 76.47 & \textbf{78.57} & 77.66 & 76.25 \\
    \midrule
    \multirow{4}{*}{MRPC}  & vanilla & 69.57 & 69.97 & 70.70 & 72.07 \\
    & \ +concat & {71.42} & 73.41 & \textbf{73.10}  & 72.80 \\
    & \ +gate & 71.01 & 72.83 & 72.97 & 72.70 \\
    & \ +cell(*) & \textbf{72.61} & \textbf{73.64} & 72.10 & \textbf{73.16} \\
    \bottomrule
    \end{tabular}%
\label{tab:transfer}%
\end{table}%

    

\subsection{Experimental Results}
Table \ref{tab:transfer} shows the accuracy results of all the models on the test sets of CR and MRPC. 
We observe that the sememe-incorporated models still outperform vanilla RNNs on both text classification datasets basically
Among the three sememe incorporation methods, +cell still performs best overall, whose average performance improvement compared with vanilla RNNs on the two datasets is $1.36$ and $2.30$, respectively.
In fact, +cell significantly outperforms vanilla models according to the results of significance test.

\section{Adversarial Attack Experiment}
Adversarial attack and defense have attracted considerable research attention recently \cite{szegedy2014intriguing,goodfellow2015explaining}, because they can disclose and fix the vulnerability of neural networks that small perturbations of input can cause significant changes of output.
Adversarial attack is aimed at generating adversarial examples to fool a neural model (victim model), and adversarial defense is targeted at improving the robustness of the model against attack.
In the field of NLP, all kinds of adversarial attack methods have been proposed \cite{zhang2019generating} but few efforts are made in adversarial defense \cite{wang2019survey}.

We intuitively believe that incorporating sememes can improve the robustness of neural networks, because sememes are general linguistic knowledge and complementary to text corpora on which neural networks heavily rely.
Therefore, we evaluate the robustness of our sememe-incorporated RNNs against adversarial attack.

\subsection{Experimental Settings}
\paragraph{Attack Method}
We use a genetic algorithm-based attack model \cite{alzantot2018generating}, which is a typical gradient-free black-box attack method. 
It generates adversarial examples by substituting words of model input iteratively and achieves impressive attack performance on both sentiment analysis and NLI.

\paragraph{Baseline Methods}
Besides vanilla RNNs, we choose adversarial training \cite{madry2017towards} as a baseline method.
Adversarial training is believed to be an effective defense method. 
It adds some generated adversarial examples to the training set, aiming to generalize the victim model to the adversarial attack.

\paragraph{Evaluation Metrics}
We test the robustness of vanilla and sememe-incorporated RNNs on SNLI. 
Robustness is measured by the attack success rate (\%), i.e., the percentage of instances in the test set which are successfully attacked by the attack model.
The lower the attack success rate is, the more robust a model is.

\paragraph{Hyper-parameter Settings}
For the attack method, we use all the recommended hyper-parameters of its original work \cite{alzantot2018generating}.
For the victim models including vanilla and sememe-incorporated RNNs, the hyper-parameters and training strategy are the same as those in previous NLI experiment.
For adversarial training, we add 57k (10\% of the number of total sentence pairs in SNLI) generated adversarial examples (premise-hypothesis pairs) to the training set.

\subsection{Experimental Results}
Table \ref{tab:attack} lists the success rates of adversarial attack against all the victim models.
We can observe that:

(1) The attack success rates of our sememe-incorporated RNNs are consistently lower than those of the vanilla RNNs, which indicates the sememe-incorporated RNNs have greater robustness against adversarial attack. 
Furthermore, the experimental results also imply the effectiveness of sememes in improving robustness of neural networks.

(2) Among the three sememe incorporation methods, +cell beats the other two once again. It demonstrates the superiority of +cell in taking full advantage of sememes.


(3) Adversarial training increases the attack success rates of all the models rather than decrease them, which is consistent with the findings of previous work \cite{alzantot2018generating}. 
It shows adversarial training is not an effective defense method, at least for the attack we use. 
In fact, there are few effective adversarial defense methods, which makes the superiority of sememes in defending adversarial attack and improving robustness of models more valuable.


\begin{table}[!tbp]
\centering
\caption{Success Rates (\%) of Adversarial Attack Against All the Victim Models on SNLI (``vanilla+at'' Represents Adversarial Training). (*) Denotes p$<$0.05 on a two-tailed t-test, against the vanilla and vanilla+at models.}
\resizebox{.9\linewidth}{!}{
    \begin{tabular}{lrrrr}
    \toprule
    Model & {LSTM} & {GRU} & {BiLSTM} & {BiGRU} \\
    \midrule
     vanilla & 65.54 & 66.30 & 64.80 & 65.46 \\
    \midrule
    vanilla+at & 66.39 & 67.55 & 65.94 & 66.60 \\
    \midrule
    +concat & 64.51 & 65.34 & 63.29 & 64.22 \\
    +gate & \textbf{62.88} & 65.14 & 62.51 & 63.63 \\
    +cell(*) & 63.98 & \textbf{65.02} & \textbf{62.40} & \textbf{63.32} \\
    \bottomrule
    \end{tabular}
}
\label{tab:attack}%
\end{table}%


\begin{table}[!t]
\centering
\caption{Language Modeling Performance of Models with Different Coverage of Sememe-annotated Words}
\resizebox{\linewidth}{!}{
    \begin{tabular}{lrrrrr}
    \toprule
     \multirow{2}{*}[-0.8ex]{Model} & \multirow{2}{*}[-0.8ex]{Coverage} & \multicolumn{2}{c}{PTB} & \multicolumn{2}{c}{WikiText-2}  \\
    \cmidrule{3-6}
     & & \multicolumn{1}{r}{LSTM} & \multicolumn{1}{r}{GRU} & \multicolumn{1}{r}{LSTM} & \multicolumn{1}{r}{GRU} \\
    \midrule
    vanilla & 0  & 80.86 & 90.68 & 93.88 & 103.04 \\
    \midrule
    \multirow{4}{*}{+concat} & 30\%   & 79.93 & 89.79 & 92.74 & 102.12 \\
     & 50\%   & 79.83 & 88.78 & 93.19 & 101.07 \\
     & 80\%   & 79.26 & 87.83 & 91.46 & 100.06 \\
     & 100\%     & \textbf{78.90}  & \textbf{87.29} & \textbf{91.41} & \textbf{98.89} \\
    \midrule
    \multirow{4}{*}{+gate}  & 30\%   & 80.08 & 88.52 & 93.00  & 100.96 \\
    & 50\%   & 79.72 & 87.90  & 92.79 & 100.15 \\
    & 80\%   & 78.62 & 86.51 & 91.47 & 98.21 \\
    & 100\%     & \textbf{77.73} & \textbf{84.35} & \textbf{90.19} & \textbf{97.54} \\
    \midrule
    \multirow{4}{*}{+cell}  & 30\%   & 80.20  & 89.97 & 93.59 & 102.02 \\
      & 50\%   & 78.76 & 88.08 & 91.89 & 101.14 \\
      & 80\%   & 78.13 & 87.74 & 90.62 & 99.23 \\
      & 100\%     & \textbf{76.65} & \textbf{86.49}& \textbf{89.16} & \textbf{97.65} \\
    \bottomrule
    \end{tabular}%
}
\label{tab:sememe-cover}%
\end{table}%

\section{Ablation Study}
To further demonstrate the effectiveness of sememes in improving sequence modeling ability of RNNs, we conduct two ablation studies in this section.
In the first study, we investigate how the model performance varies with the coverage of sememe-annotated words.
In the second one, we substitute sememes with other information to show the superiority of sememes.

\subsection{Sememe Coverage Experiment}
In this experiment, we deliberately drop the sememe annotations of a certain percent of annotated words and then re-train and evaluate the sememe-incorporated models on PTB and WikiText-2.
Table \ref{tab:sememe-cover} lists the perplexity results of the medium models with different coverage of sememe-annotated words (vanilla or 0, 30\%, 50\%, 80\% and 100\%) on the test sets of PTB and WikiText-2. 
We can clearly see that with the increase of the coverage of sememe-annotated words, performance of all models on whichever dataset becomes higher. 
Considering the models with different coverage of sememe-annotated words have the same number of parameters (except the vanilla RNNs), these results can convincingly demonstrate the effectiveness of sememes in improving the sequence modeling ability of RNNs.

\subsection{Sememe Substitution Experiment}
In this experiment, we try to substitute sememes with other information to improve RNNs.
First, we randomly assign each word some \textit{meaningless labels} in the same quantity as its sememes, where the total number of different labels is also equal to that of sememes.
In addition, we use WordNet \cite{miller1998wordnet} as a source of external knowledge and substitute the sememes of a word by its synonyms with the same POS tag.
We use the three sememe incorporation methods to incorporate above information into different RNNs and evaluate corresponding models on SNLI.
Experimental results are shown in Table \ref{tab:snli_ablation}.
We can find that for the same knowledge incorporation method, models incorporated with sememes obviously outperform those incorporated with meaningless labels or WordNet synonyms.
These results manifest the superiority of sememes.

\begin{table}[!tbp]
  \small
  \centering
     \caption{Performance of Models Incorporated with Different Information on SNLI}
    \begin{tabular}{clllll}
    \toprule
    Information & Model & {LSTM} & {GRU} & {BiLSTM} & {BiGRU} \\
    \midrule
    None & vanilla & 81.03 & 81.59 & 81.38 & 81.92 \\
    \midrule
    \multirow{3}{*}{\shortstack{ Meaningless \\ Label}} & +concat & 80.81 & 81.37 & 82.13 & 82.16 \\
    & +gate & 79.37 & 80.93 & 80.84 & 79.35 \\
    & +cell & 78.92 & 81.52 & 81.78 & 81.24 \\
    \midrule
    \multirow{3}{*}{WordNet} & +concat & 80.19 & 81.36 & 82.14 & 82.37 \\
    & +gate & 80.97 & 81.78 & 82.45 & 81.68 \\
    & +cell & 81.42 & 81.75 & 82.33 & 81.79 \\
    \midrule
    \multirow{3}{*}{Sememe} & +concat & \textbf{81.72} & 81.87 & 82.39  & 82.70 \\
    & +gate & 81.61 & 82.33 & 83.06 & 83.18 \\
    & +cell & 81.66 & \textbf{82.78} & \textbf{83.67} & \textbf{83.26} \\
    \bottomrule
    \end{tabular}%
  \label{tab:snli_ablation}%
\end{table}%

\section{Related Work}



\subsection{HowNet and Its Applications}
HowNet \cite{dong2003hownet} is one of the most famous sememe KBs, whose construction takes several linguistic experts more than two decades. 
After HowNet is published, it has been employed in diverse NLP tasks including word similarity computation \cite{liu2002word}, word sense disambiguation \cite{duan2007word}, word representation learning \cite{niu2017improved}, sentiment analysis \cite{fu2013multi}, etc.
Recently, with the development of deep learning, sememe knowledge in HowNet has also been incorporated into neural models to improve performance.
\cite{qi2019modeling} consider sememes as the external knowledge of semantic composition modeling and obtain better representations of multi-word expressions.
\cite{zang2019textual} utilize HowNet to find candidate substitute words in word-level textual adversarial attacks and achieve higher attack performance as compared with other word substitution methods.
\cite{zhang2020multi} regard sememes as the semantic features of words and realize a more effective and robust reverse dictionary model.

Among the above researches, \cite{gu2018language} exploit sememes in a sequence modeling task (language modeling) for the first time. 
They add a sememe predictor to an LSTM language model, which is aimed at predicting sememes of the next word using preceding text.
Then they use the predicted sememes to predict the next word.
Their method uses sememes in the decoder step of the LSTM language model and does not improve the LSTM's ability to encode sequences.
Therefore, it cannot be applied to other sequence modeling tasks.
As far as we know, we are the first to utilize sememes to improve general sequence modeling ability of neural networks.

Another line of researches about HowNet is automatic construction and updating of sememe KBs, mostly by lexical sememe prediction.
\cite{xie2017lexical} present the task of sememe prediction and propose two simple but effective sememe prediction methods.
\cite{jin2018incorporating} incorporate Chinese character information in sememe prediction and obtain higher performance.
To automatically build a sememe KB like HowNet for other languages, \cite{Qi2018Cross} propose to make cross-lingual lexical sememe prediction for unlabeled words in a new language. 
\cite{qi2020towards} go further and try to build a multilingual sememe KB for multiple languages based on a multilingual encyclopedic dictionary.

\subsection{Recurrent Neural Networks }
The recurrent neural network (RNN) \cite{rumelhart1988learning} and its representative extentions including LSTM \cite{hochreiter1997long} and GRU \cite{cho2014learning}, have been widely employed in various NLP tasks, e.g., language modeling \cite{mikolov2010recurrent}, sentiment analysis \cite{nakov2016semeval}, semantic role labelling \cite{he2018jointly}, dependency parsing \cite{kiperwasser2016simple} and natural language inference \cite{parikh2016decomposable}.

To improve the sequence modeling ability of RNNs, some researches try to reform the frameworks of RNNs, e.g., integrating the attention mechanism \cite{bachman2015variational}, adding hierarchical structures \cite{schmidhuber1992learning} and introducing bidirectional modeling \cite{graves2005bidirectional}. 

In addition, some work focuses on incorporating different kinds of external knowledge into RNNs. 
General linguistic knowledge from famous KBs such as WordNet \cite{miller1998wordnet} and ConceptNet \cite{speer2013conceptnet} attracts considerable attention. 
These KBs usually comprise relations between words, which are hard to be incorporated into the internal structures of RNNs.
Therefore, most existing knowledge-incorporated methods employ external linguistic knowledge on the hidden layers of RNNs rather than the internal structures of RNN cells
\cite{ahn2016neural,yang2017leveraging,parthasarathi2018extending,young2018augmenting}.

Since sememe knowledge is very different from the word-level relational knowledge, it cannot be incorporated into RNNs with the same methods. 
As far as we know, no previous work tries to incorporate such knowledge as sememes into RNNs.



\section{Conclusion and Future Work}
In this paper, we make the first attempt to incorporate sememes into RNNs to enhance their sequence modeling ability, which is beneficial to many downstream NLP tasks.
We preliminarily propose three highly adaptable sememe incorporation methods, and employ them in typical RNNs including LSTM, GRU and their bidirectional versions. 
In experiments, we evaluate our methods on several representative sequence modeling tasks.
Experimental results show that sememe-incorporated RNNs achieve obvious performance improvement, which demonstrates the usefulness of sememes and effectiveness of our methods.

  
In the future, we will explore following directions including:
(1) considering the hierarchical structures of sememes, which contain more semantic information of words; 
(2) using attention mechanism to adjust the weights of sememes in different context to take better advantage of sememes; 
(3) evaluating our sememe-incorporated RNNs on other sequence modeling tasks; 
and (4) incorporating sememe knowledge into other neural models including the tailor-made RNNs and the Transformer.

%
%


%

\bibliographystyle{IEEEtran}
\bibliography{ref}

%

\begin{IEEEbiography}[{\includegraphics[width=1in,height=1.25in,clip,keepaspectratio]{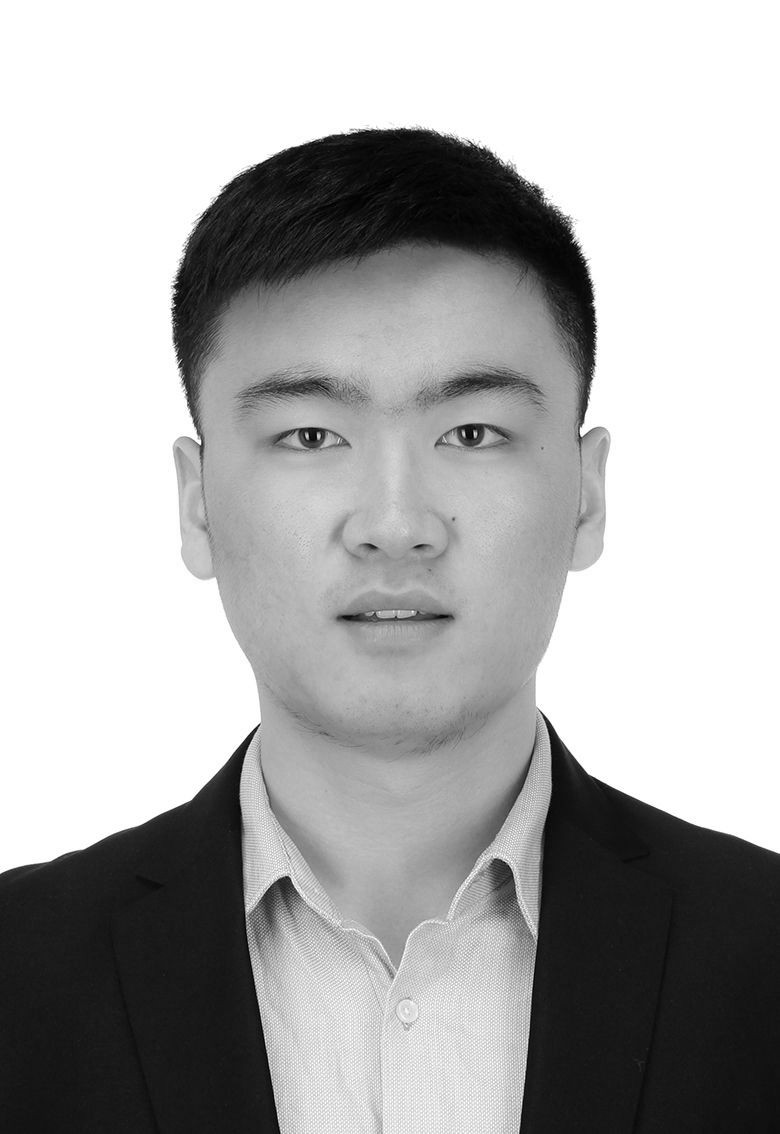}}]{Yujia Qin}
received the B.E. degree from the Department of Electronic Engineering, Tsinghua University, Beijing, China, in 2020. He is currently working toward the Ph.D. degree with the Department of Computer Science, University of Illinois at Urbana-Champaign. His research interests include natural language processing and machine learning.
\end{IEEEbiography}

\begin{IEEEbiography}[{\includegraphics[width=1in,height=1.25in,clip,keepaspectratio]{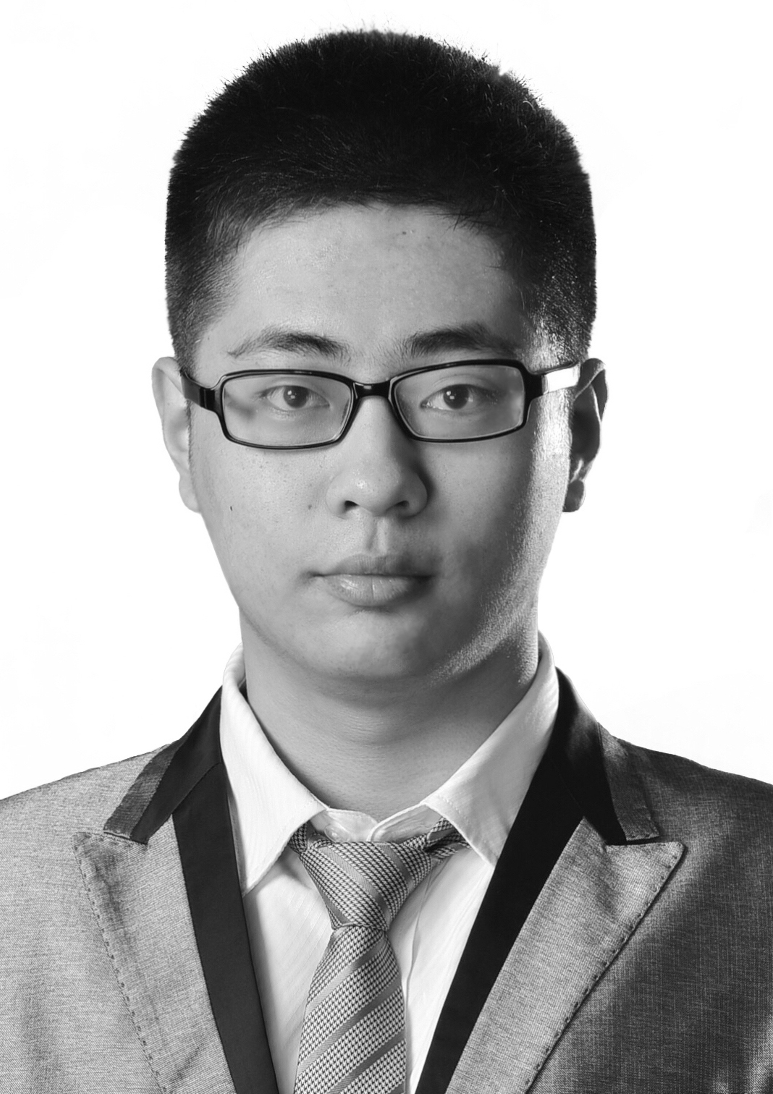}}]{Fanchao Qi}
received the B.E. degree from the Department of Electronic Engineering, Tsinghua University, Beijing, China, in 2017. He is currently working toward the Ph.D. degree with the Department of Computer Science and Technology, Tsinghua University, Beijing, China. His research interests include natural language processing and computational semantics.
\end{IEEEbiography}


\begin{IEEEbiography}[{\includegraphics[width=1in,height=1.25in,clip,keepaspectratio]{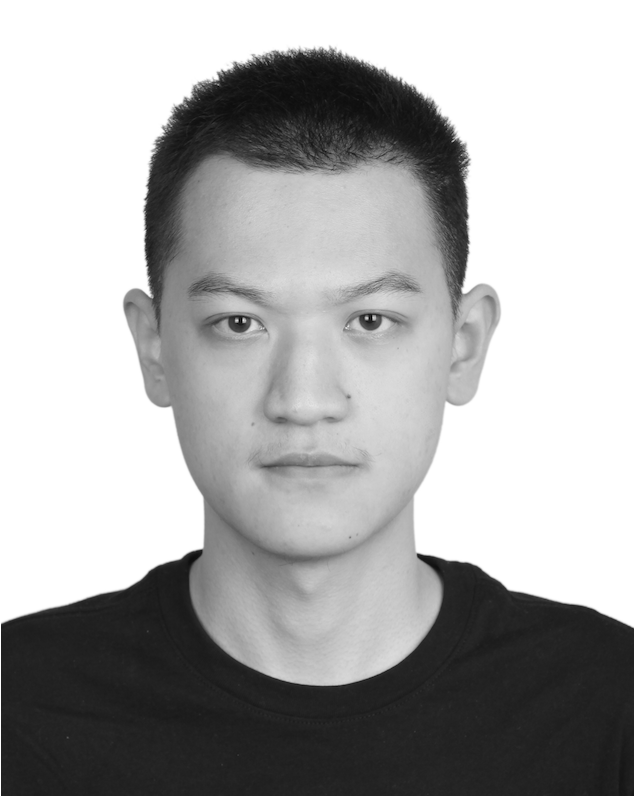}}]{Sicong Ouyang}
received the B.E. degree from Beijing University of Posts and Telecommunications, Beijing, China, in 2020. He is now working at a private equity. His research interests include natural language processing and graph representation learning.
\end{IEEEbiography}

\begin{IEEEbiography}[{\includegraphics[width=1in,height=1.25in,clip,keepaspectratio]{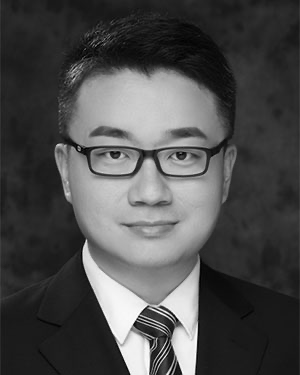}}]{Zhiyuan Liu}
received the Ph.D. degree from the Department of Computer Science and Technology, Tsinghua University, Beijing, China, in 2011. He is currently an associate professor with the Department of Computer Science and Technology, Tsinghua University, Beijing, China. His research interests include natural language processing, knowledge graph and social computation.
\end{IEEEbiography}

\vfill
\newpage
\begin{IEEEbiography}[{\includegraphics[width=1in,height=1.25in,clip,keepaspectratio]{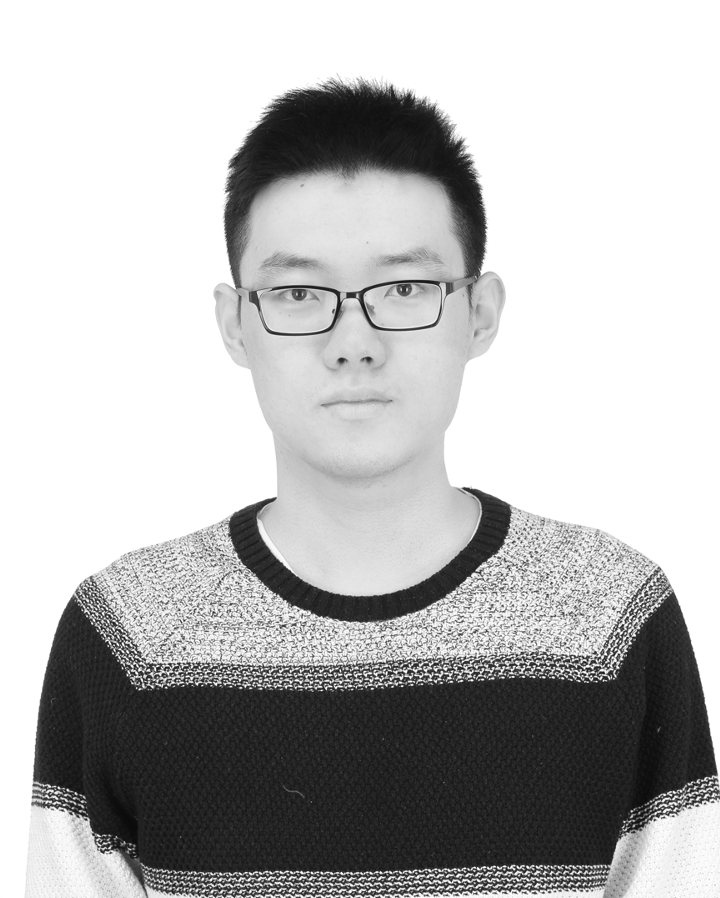}}]{Cheng Yang}
received the Ph.D. degree rom the Department of Computer Science and Technology, Tsinghua University, Beijing, China, in 2019. He is currently an assistant professor at Beijing University of Posts and Telecommunications, Beijing, China. His research interests include natural language processing and network representation learning.
\end{IEEEbiography}

\begin{IEEEbiography}[{\includegraphics[width=1in,height=1.25in,clip,keepaspectratio]{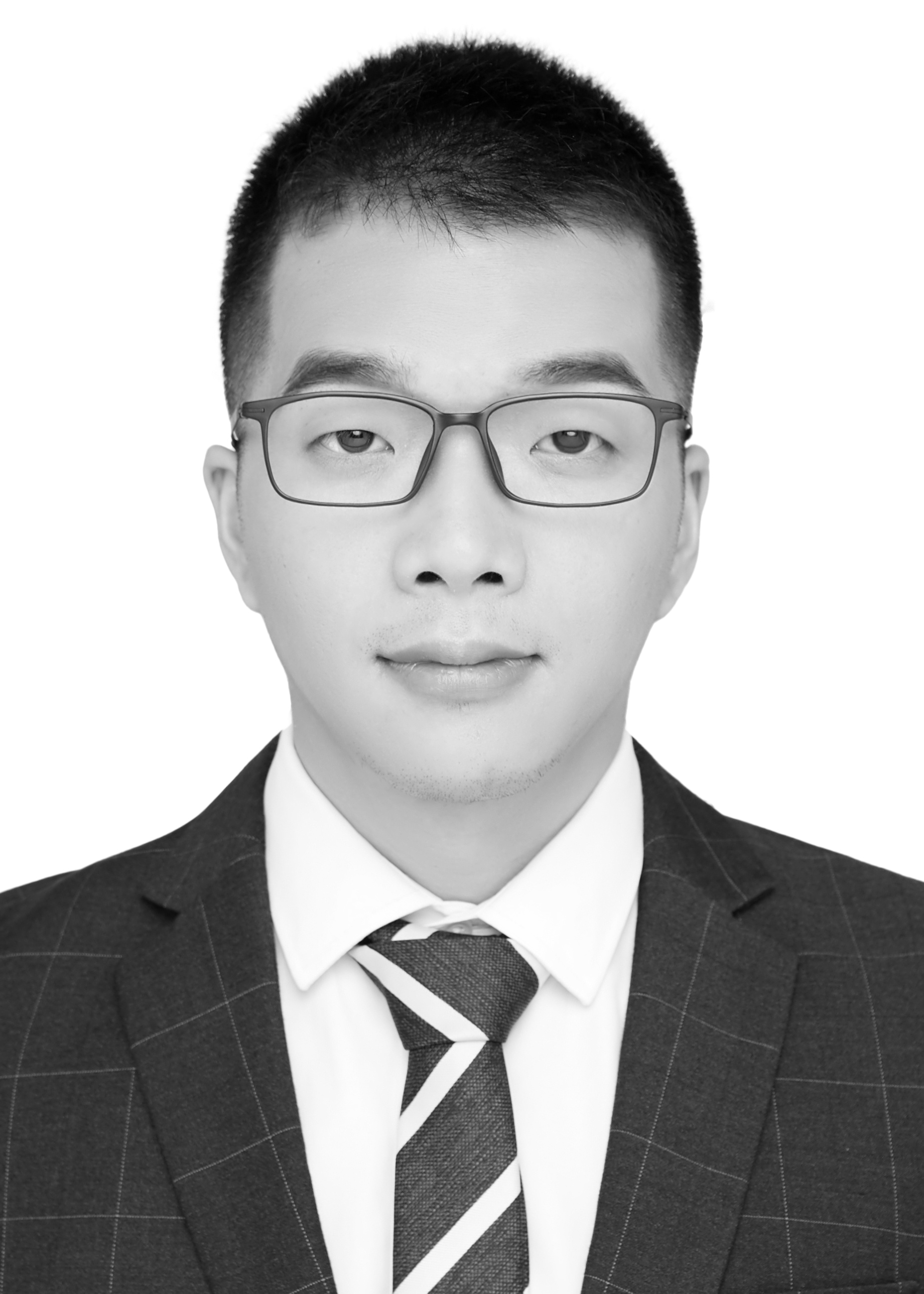}}]{Yasheng Wang}
received the Ph.D. degree from Zhejiang University, Hangzhou, China, in 2015. He is currently a researcher of Speech and Language Computing of Huawei Noah's Ark Lab. His research interests include natural language processing and dialog system.
\end{IEEEbiography}

\begin{IEEEbiography}[{\includegraphics[width=1in,height=1.25in,clip,keepaspectratio]{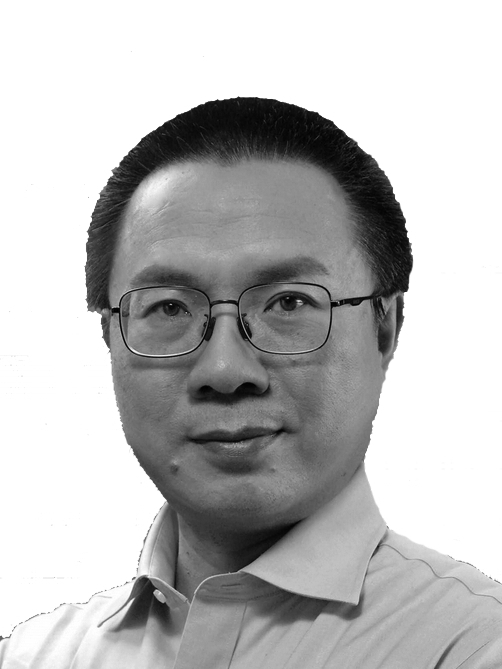}}]{Qun Liu}
received the Ph.D. degree from Peking University, Beijing, China, in 2004. He is currently the Chief Scientist of Speech and Language Computing of Huawei Noah's Ark Lab. His research interests include natural language processing, machine translation, dialog system, etc.
\end{IEEEbiography}

\begin{IEEEbiography}[{\includegraphics[width=1in,height=1.25in,clip,keepaspectratio]{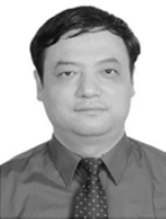}}]{Maosong Sun}
received the Ph.D. degree in computational linguistics from City University of Hong Kong, Hong Kong, in 2004. He is currently a professor with the Department of Computer Science and Technology, Tsinghua University, Beijing, China. His research interests include natural language processing, web intelligence, and machine learning.
\end{IEEEbiography}
\vfill



\end{document}